\documentclass{bmvc2k}

\usepackage{dsfont}
\usepackage{nicefrac}

\title{Unsupervised Point Cloud Registration with Self-Distillation}

\addauthor{Christian L\"owens}{christian.loewens@bosch.com}{1,2}
\addauthor{Thorben Funke}{thorben.funke@bosch.com}{1}
\addauthor{Andr\'e Wagner}{andre.wagner2@bosch.com}{1}
\addauthor{Alexandru Paul Condurache}{alexandrupaul.condurache@bosch.com}{2,3}
\author{Christian L\"owens \AND Thorben Funke \AND Andr\'e Wagner \AND Alexandru Paul Condurache}
\hypersetup{
    pdftitle={Unsupervised Point Cloud Registration with Self-Distillation},
    pdfauthor={Christian L\"owens, Thorben Funke, Andr\'e Wagner, Alexandru Paul Condurache}
}

\addinstitution{
 Bosch Research
}
\addinstitution{
 University of L\"ubeck
}
\addinstitution{
 Robert Bosch GmbH\\ Automated Driving
}

\runninghead{L\"owens et al.}{Unsupervised Registration with Self-Distillation}

\def\etal{\emph{et al}\bmvaOneDot}

\newcommand{\ours}{DiReg}

\DeclareMathOperator*{\argmin}{arg\,min}

\newcommand{\real}{\mathbb{R}}

\newcommand{\transf}{{T}}
\newcommand{\rot}{{R}}
\newcommand{\transl}{{t}}
\newcommand{\corr}[1][]{\mathcal{C}_{#1}}
\newcommand{\feat}[1][]{\mathcal{F}_{#1}}
\newcommand{\pointA}[1][]{\mathcal{A}_{#1}}
\newcommand{\pointB}[1][]{\mathcal{B}_{#1}}
\newcommand{\augpointA}{\widetilde{\pointA}}
\newcommand{\augpointB}{\widetilde{\pointB}}

\newcommand{\numFeatures}{k}
\newcommand{\numLatentFeatures}{\ell}
\newcommand{\pointP}[1][]{\mathcal{P}_{#1}}

\newcommand{\corrEst}{{\corr}_{\text{raw}}}
\newcommand{\corrEstNew}{{\corr}_{\text{ref}}}

\begin{document}

\maketitle
\begin{abstract}
Rigid point cloud registration is a fundamental problem and highly relevant in robotics and autonomous driving. Nowadays deep learning methods can be trained to match a pair of point clouds, given the transformation between them. However, this training is often not scalable due to the high cost of collecting ground truth poses. Therefore, we present a self-distillation approach to learn point cloud registration in an unsupervised fashion. Here, each sample is passed to a teacher network and an augmented view is passed to a student network. The teacher includes a trainable feature extractor and a learning-free robust solver such as RANSAC. The solver forces consistency among correspondences and optimizes for the unsupervised inlier ratio, eliminating the need for ground truth labels. Our approach simplifies the training procedure by removing the need for initial hand-crafted features or consecutive point cloud frames as seen in related methods. We show that our method not only surpasses them on the RGB-D benchmark 3DMatch but also generalizes well to automotive radar, where classical features adopted by others fail. The code is available at \href{https://github.com/boschresearch/direg}{github.com/boschresearch/direg}.
\end{abstract}

\section{Introduction}
\label{sec:intro}

\begin{figure}[t]
    \centering
    \includegraphics[width=\textwidth]{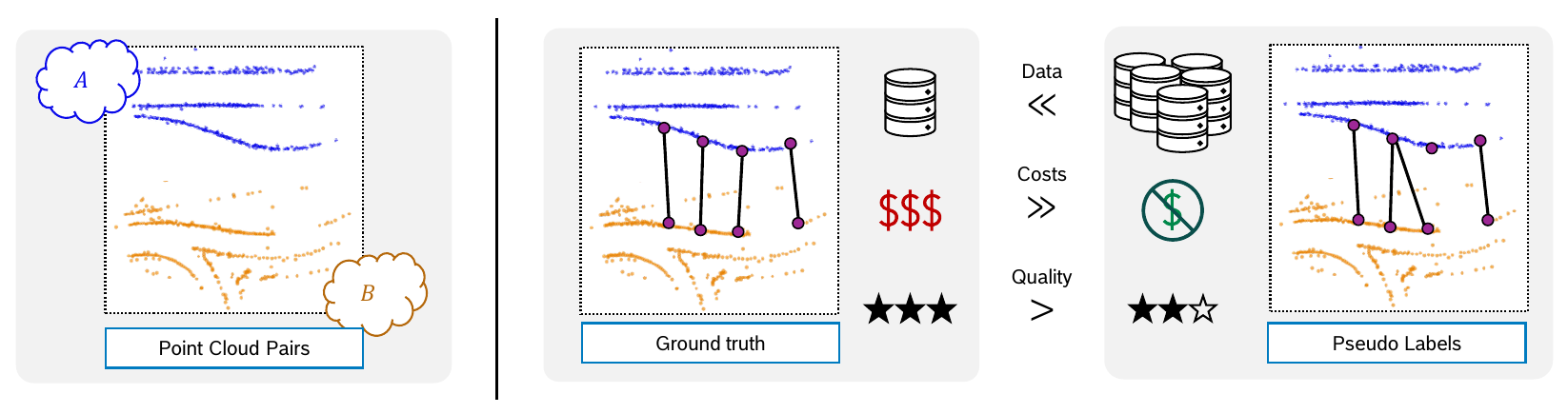}
    \caption{\textbf{Motivation for Unsupervised Point Cloud Registration.} Especially in the automotive context, the collection of ground truth poses is costly and limited in size. Crowdsourced data from consumer-grade cars, on the other hand, contains orders of magnitude more unlabeled data. Using our approach, we can leverage this data and generate pseudo labels with a quality close to ground truth.}
    \label{fig:motivation}
\end{figure}

The goal of rigid point cloud registration is to align two or more point clouds by finding the optimal rigid transformation. It is a fundamental task in fields such as 3D reconstruction \cite{Choi_2015_CVPR, gojcic2020learning}, augmented reality \cite{info14030149, mahmood20193d} and autonomous navigation \cite{electronics8010043, zhang2014loam}.
Traditionally, these applications have relied on learning-free heuristics \cite{fischler1981random, besl1992method, rusu2009fast} or supervised deep learning approaches \cite{zeng20173dmatch, choy2019fully, hisadome2021cascading} that require ground truth poses during training.
While these methods are effective, they often lack scalability across diverse scenarios.
In the automotive context, ground truth surveys are expensive and limited in size due to their professional sensor setup \cite{Geiger2013IJRR}. However, crowdsourced data \cite{10171417} can be obtained on a large scale from mass-produced vehicles, providing more diverse data as shown in Fig.~\ref{fig:motivation}. Unsupervised point cloud registration can utilize this and generate high-quality pseudo labels at a low cost.

In recent years, some methods have emerged to overcome the need for ground truth \cite{liu2023self, yang2021self, shen2022reliable}. Notably, Yang \etal \cite{yang2021self} draw inspiration from student-teacher architectures and propose Self-supervised Geometric Perception (SGP). There, the student is a trainable feature matcher outputting putative correspondences and the teacher is a learning-free robust solver estimating rigid transformations (see Fig.~\ref{fig:sgp}). The transformations are then used as pseudo labels to supervise the student for several epochs before new improved labels are generated again. Revisiting SGP, we adopt some of the research findings in self-distillation, where student and teacher are both parameterized feature extractors, leading to remarkable unsupervised performance in the image domain \cite{grill2020bootstrap, caron2021emerging}.
For this, we update the teacher by an exponential moving average (EMA) of the student's parameters to provide continuously better pseudo labels on the fly.
Furthermore, we propose a more simplified framework compared to SGP (see Fig.~\ref{fig:direg}) by eliminating the pseudo label verifier and the reliance on hand-crafted bootstrap features. This enhances the adaptability of our method to various modalities, whereas the previously mentioned SGP components may require careful adjustments or may not work at all.
In this process, we converge on an architecture similar to the most recently published work Extend Your Own Correspondences (EYOC)~\cite{liu2024extend}. However, EYOC focuses specifically on distant LiDAR point cloud registration, while we investigate the general unsupervised problem in diverse environments. Moreover, we show that the augmentation technique used in EYOC and others makes it difficult to robustly bootstrap the training process, and thus remove the augmentation for the teacher's input.

Our method surpasses the performance of both SGP and EYOC on 3DMatch and a radar dataset while having fewer hyperparameters requiring tuning.
In summary, we propose a self-\textbf{di}stillation approach for \textbf{reg}istration (DiReg) with the following key contributions:
\begin{itemize}
    \item We simplify unsupervised point cloud registration by removing the need for a pseudo label verifier, hand-crafted bootstrap features, and progressive datasets as seen in earlier work, which all need careful adjustments when used for data from various sensors.
    \item We demonstrate that the data augmentation commonly used for the teacher's input can impede robust bootstrapping in unsupervised settings and how to overcome it.
    \item We show that our approach generalizes well across modalities as evidenced by its performance on RGB-D and automotive radar point clouds.
\end{itemize}

\section{Related Work}
\label{sec:related}

\subsection{Point Cloud Registration}
In point cloud registration, we align two point clouds either directly or estimate a set of correspondences first. Usually, those correspondences are computed by trained feature matchers and then passed to a robust estimator filtering out  the outliers to predict the transformation.

\noindent\textbf{Feature Matchers:}
Prior to learning-based methods, Fast Point Feature Histograms (FPFH) \cite{rusu2009fast} uses hand-crafted features to capture the local geometry.  Fully Convolutional Geometric Features (FCGF) \cite{choy2019fully} introduces sparse 3D convolutions to learn the feature extraction. GeoTransformer \cite{qin2022geometric} integrates self- and cross-attention and estimates correspondences by computing the optimal transport, which PEAL \cite{yu2023peal} improves by modeling unidirectional attention from putative non-overlapping to overlapping superpoints. Recently, BUFFER \cite{ao2023buffer} combines patch-wise and point-wise methods to improve generalizability. All learning-based methods need ground truth correspondences during training, which is usually expensive to obtain on a large scale.

\noindent\textbf{Robust Estimators:}
Robust estimators optimize for the best transformation given the putative correspondences by the feature matcher. The Random Sample Consensus (RANSAC) \cite{fischler1981random} algorithm estimates transformations on random correspondence subsets and is widely utilized due to its robustness against a high percentage of outliers. The Iterative Closest Point (ICP) \cite{besl1992method} method iteratively adjusts an initial transformation and the correspondence pairs. SC$^2$-PCR \cite{chen2022sc2} searches for a spatial compatibility consensus among correspondences to better distinguish inliers and outliers. Recent methods like $\nabla$RANSAC \cite{wei2023generalized} allow fully-differentiable pipelines to train feature extractors in an end-to-end fashion.

\begin{figure}[t]
     \centering
     \begin{subfigure}[c]{0.58\textwidth}
         \centering
         \includegraphics[width=\textwidth]{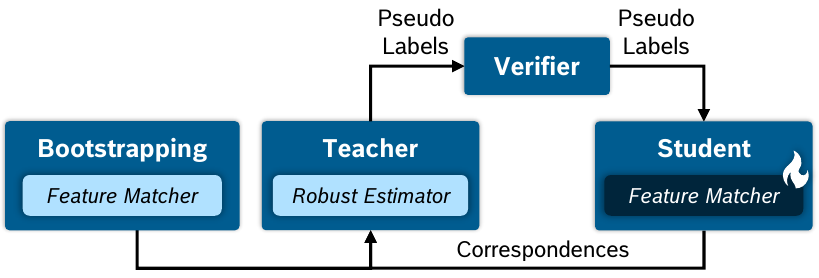}
         \caption{SGP \cite{yang2021self}}
         \label{fig:sgp}
     \end{subfigure}
     \hfill
     \begin{subfigure}[c]{0.38\textwidth}
         \centering
         \vspace{9pt}
         \includegraphics[width=\textwidth]{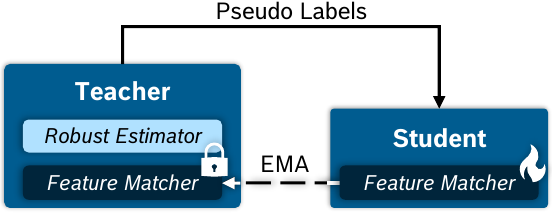}
         \vspace{0pt}
        \caption{\ours\ (Ours)}
         \label{fig:direg}
     \end{subfigure}
     \hfill
     \vspace{10pt}
        \caption{We simplify the SGP algorithm by removing the verifier and its classical features used for bootstrapping. We also reinterpret its student-teacher analogy in view of self-distillation. While dark boxes indicate trainable methods, the teacher's feature extractor is not trained but instead updated using an exponential mean average (EMA).}
        \label{fig:sgp_vs_direg}
\end{figure}

\noindent\textbf{Unsupervised Registration:}
Despite the prevalence of supervised feature extraction, some studies have demonstrated the feasibility of unsupervised training. SGP \cite{yang2021self} draws an analogy to student-teacher models as illustrated in Fig.~\ref{fig:sgp}. In this context, the student is a trainable feature extractor training on the same pseudo labels for a number of epochs until the teacher, which is RANSAC followed by ICP, generates new labels. For the initial pseudo labels, FPFH features are used. EYOC \cite{liu2024extend} extends this idea to automotive datasets, but its teacher incorporates a momentum encoder and SC$^2$-PCR as the learning-free solver. It utilizes consecutive frames from LiDAR sequences and progressively learns to register point clouds that are more distant from each other. Moreover, they spatially filter correspondences close to the ego vehicle. BYOC \cite{el2021bootstrap} exploits the fact that images and point clouds are coupled in RGB-D data and trains a point cloud feature extractor with pseudo labels coming from a randomly initialized image feature extractor. UDPReg \cite{mei2023unsupervised} models point clouds as GMMs and leverages consistency in feature and coordinate space as a self-supervisory signal, while RIENet \cite{shen2022reliable} learns a neighborhood consensus between correspondences.

\subsection{Self-Distillation}
Several methods in the image domain demonstrate the effective use of the model's own predictions to enhance learning without extensive labeled data. The Mean Teacher model \cite{tarvainen2017mean} relies on a student-teacher architecture, where the teacher is an EMA of the student's parameters, fostering consistency in predictions for semi-supervised learning. BYOL \cite{grill2020bootstrap} uses a mean teacher to remove the need for labels completely. Caron \etal \cite{caron2021emerging} propose self-distillation with no labels (DINO) by changing the non-contrastive loss in BYOL to a cross-entropy loss on pseudo-class labels. Xie \etal \cite{xie2020self} introduce noise and data augmentation into the student's training, aiming to replicate the teacher's output, improving generalization.

\section{Problem Formulation}
Given two partially overlapping 3D point clouds $\pointA\subset\real^{3}$ and $\pointB\subset\real^{3}$, we want to find the optimal rigid transformation $\transf^*=\{\rot^*\in \operatorname{SO}(3), \transl^*\in\real^3\}$ for a set of  ground truth correspondences $\mathcal{C}^*$ with a minimal mean-squared error ($\operatorname{MSE}$):
\begin{equation}
    \label{eq:align}
    \transf^* = \argmin_{{\transf}}~ \operatorname{MSE}\left(\mathcal{C}^*, {\transf}\right) = \argmin_{{\transf}=\{{\rot}, {\transl}\}}{\sum_{(a,b)\in\mathcal{C}^*}{||{\rot}a+{\transl}-b||_2^2}}
\end{equation}
Here,  $a,b\in\mathbb{R}^{3}$ denote the coordinates of two corresponding points in $\pointA$ and $\mathcal{B}$. However, ground truth labels are usually not given and have to be estimated first. Depending on the dataset, both point clouds might hold $\numFeatures$ additional features such as color. For simplicity, we denote the combination of coordinates and features as $\pointA\subset\real^{3+\numFeatures}$ and $\pointB\subset\real^{3+\numFeatures}$. Feature matchers can learn a function $m_\theta$ to estimate the correspondences with $\corr = m_{\theta}({\pointA},{\pointB})$, but usually require the ground truth pose $\transf^*$ during the training process. We tackle the harder problem where no ground truth is available.

\section{Method}
\begin{figure}
    \centering
    \includegraphics[width=\textwidth]{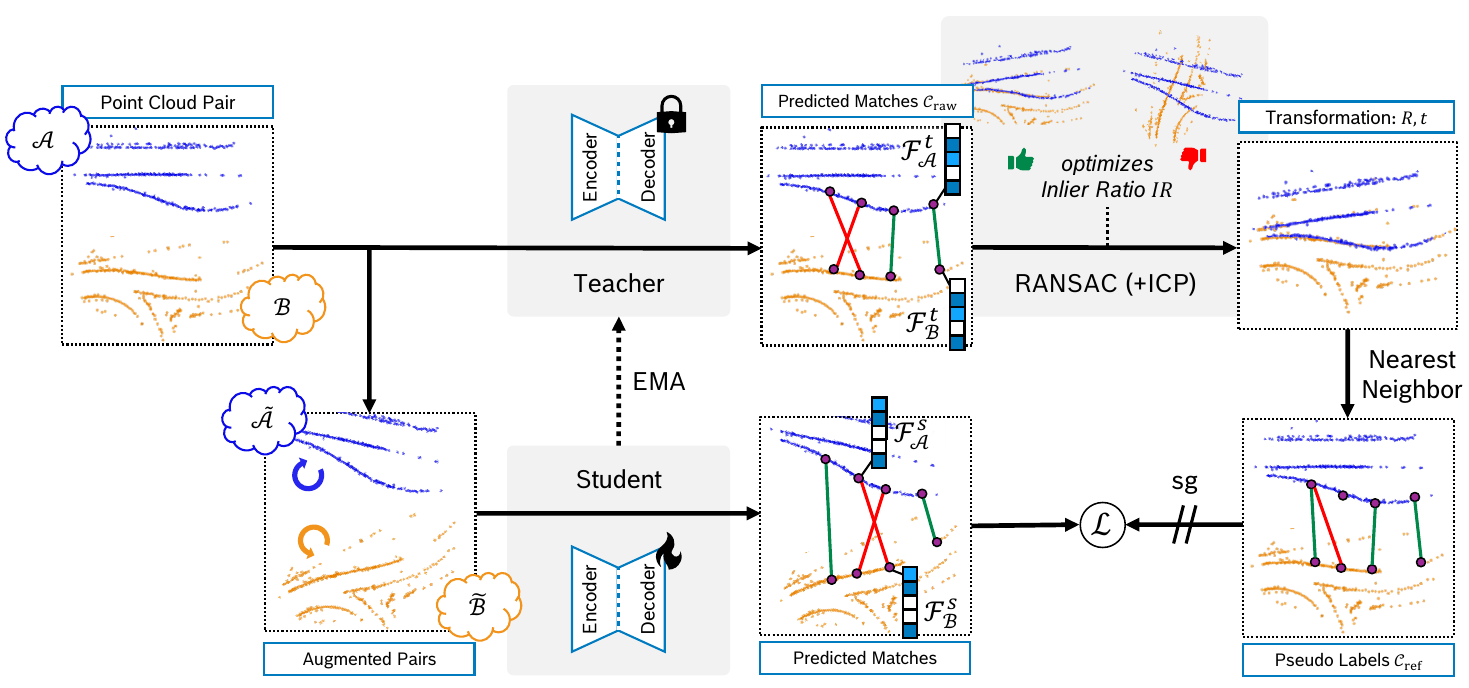}
\caption{\textbf{Self-distillation for registration (DiReg).} Both point clouds are passed to the teacher, while the student receives the augmented views. The networks, FCGF \cite{choy2019fully} feature extractors, predict geometric features for all points in their pairs and we collect correspondences by searching for the nearest neighbors among the feature vectors of the teacher. Given those correspondences, RANSAC estimates a transformation to align both clouds. Next, we search for nearest neighbors in the coordinate space to get improved correspondences for supervising the student. sg denotes the stop-gradient operator to illustrate that we do not backpropagate through the teacher network. Best viewed on display.}
    \label{fig:process}
\end{figure}

We adopt a student-teacher architecture, where the teacher generates pseudo labels on the fly to train the student. This continuously improves the pseudo labels, in contrast to SGP, which generates new labels only after several epochs of training. Our teacher network is updated using an EMA of the student's parameters and therefore, shares the same architecture. Since we are in an unsupervised setting, this process is also termed self-distillation \cite{caron2021emerging}. We incorporate a robust solver into our teacher to improve its estimation. In the final step, we apply a contrastive loss, where the positive pairs are determined by the teacher's estimated correspondences.
Fig.~\ref{fig:process} visualizes the training process.

\subsection{Feature Matching} 
The feature matcher consists of the commonly used FCGF \cite{choy2019fully} feature extractor and a subsequent nearest neighbor search. This allows for a more accurate comparison of our method with existing unsupervised approaches  \cite{yang2021self,liu2024extend} by eliminating any potential bias due to differences in the backbone network. In the forward pass, both featured point clouds  ${\pointP}\in\{{\pointA}, {\pointB}\}$ are voxelized and passed to the teacher's feature extractor, a 3D variant of a ResUNet with sparse convolutions \cite{choy2019fully}. It predicts latent features $\mathcal{F}_\mathcal{P}\subset\real^{\numLatentFeatures}$ for all points in ${\pointP}$. The student receives an augmented view of both point clouds denoted as  $\augpointA$ and $\augpointB$. We discuss this design choice in more detail in section \ref{sec:bootstrapping}. The extracted features of the student and the teacher will be denoted as $\feat[\pointP]^s$  and $\feat[\pointP]^t$.

Then we search for all points in $\augpointA$ the corresponding points in $\augpointB$, where the distance between the features of our teacher $\mathcal{F}^t_{\pointA}$ and $\mathcal{F}^t_{\pointB}$  is minimal, resulting in the initially estimated correspondences $\corrEst=\operatorname{NN}(\mathcal{F}^t_{\pointA}, \mathcal{F}^t_{\pointB})$.

\subsection{Pseudo label generation}
Directly using the correspondences $\corrEst$ from an untrained feature matcher as pseudo labels results in unsatisfactory performance \cite{el2021bootstrap}.
Therefore, we improve the correspondence prediction by adopting RANSAC optionally followed by the ICP algorithm as proposed in SGP \cite{yang2021self}. Both solvers optimize $\hat{\transf}=\{\hat{\rot},\hat{\transl}\}$ with respect to the unsupervised inlier ratio ($\operatorname{IR}$) defined as:
\begin{equation}\label{eq:inlier}
    \operatorname{IR}(\hat{\transf}) =\frac{1}{|\corrEst|}{\sum_{(a,b)\in\corrEst}{\mathds{1}\left[||\hat{\rot}a+\hat{\transl}-b||<\tau_1\right]}},
\end{equation}
where $\mathds{1}[\cdot]$  is the indicator function, $||\cdot||$ is the L$_2$ norm and $\tau_1$ is the acceptable distance threshold. In contrast to the weighted Procrustes solver used in BYOC \cite{el2021bootstrap}, RANSAC is more robust to outliers and is therefore widely used. However, since the original RANSAC method is not differentiable, this choice prevents us from directly backpropagating a pose loss as in BYOC and limits us to using a correspondence-based loss with pseudo labels. To generate the new refined correspondences $\corrEstNew$ from the optimized transformation $\hat{\transf}$, we perform a nearest neighbor search in the coordinate space with $\corrEstNew=\operatorname{NN}(\pointA\hat{\rot}^\top+\hat{\transl},\pointB)$ and keep only those with a distance below a second threshold $\tau_2$.

Additionally, SGP proposed a verifier to remove samples with transformations, where the inlier ratio is below a certain threshold. While this slightly improves the training runtime, we consistently saw a decrease in performance (see Section~\ref{sec:ablation}) and therefore did not include it.

\subsection{Unsupervised Training}
\textbf{Loss Function:} We adapt the hardest-contrastive loss $\mathcal{L}$ \cite{choy2019fully} for training the student network. It combines the loss $\mathcal{L}^p$ for the positive pairs with the hardest negative losses  $\mathcal{L}^n_{\pointA\pointB}$ and  $\mathcal{L}^n_{\pointB\pointA}$ for the first and the second element in each pair.
\begin{equation}
    \mathcal{L}(\corrEstNew, \feat[\pointA]^{s}, \feat[\pointB]^{s}) = \mathcal{L}^p + \nicefrac{1}{2}\left(\mathcal{L}^n_{\pointA\pointB} 
    + \mathcal{L}^n_{\pointB\pointA}\right)
\end{equation}
Thereby, we force the features for positive pairs to be close and for negative pairs to be distant. Since no ground truth labels are available, our positive pairs are the correspondences predicted by the teacher $\corrEstNew$, and negative pairs are determined accordingly. Note that we do not backpropagate gradients through the teacher.

\noindent\textbf{Update-Strategy:} We update the teacher's parameters $\theta^t_i$ at each step $i$ with an exponential moving average (EMA) of the student's parameters $\theta^s$ as proposed by Mean Teacher \cite{tarvainen2017mean}:
\begin{equation}
    \label{eq:update}
    \theta^t_i = \alpha\theta^t_{i-1} + (1-\alpha)\theta^s_{i},
\end{equation}
where $\alpha$ follows a cosine schedule \cite{grill2020bootstrap} from 0.9 to 1. Empirically, we get slightly better results compared to an architecture, where student and teacher share the same parameters, i.e. $\alpha=0$.
Moreover, this update strategy leads to a continuous improvement of our pseudo labels in contrast to SGP, where new labels are only generated after a complete training run.

\subsection{Data Augmentation and Bootstrapping}
\label{sec:bootstrapping}

We follow the augmentation for FCGF \cite{choy2019fully} and randomly rotate both point clouds to force the network to become rotation invariant. However, in our distillation process, it is important to overcome the bootstrap phase, where the randomly initialized teacher can hardly provide any beneficial pseudo labels. Here, the teacher is not yet trained to be invariant against rotations and thus performs weakly with augmented samples. To mitigate this, we draw inspiration from Noisy Student Training \cite{xie2020self} and only apply the augmentation on the student's input. Surprisingly, keeping the data's original orientation substantially impacts the bootstrap phase as we demonstrate in our ablations (see Section~\ref{sec:ablation}). It makes the registration task much easier and accelerates the training. Nevertheless, we maintain the augmentation for the student so that our model learns the same invariance. Note that this approach is different from EYOC \cite{liu2024extend}, which does the augmentation for the student and the teacher network.

SGP utilizes classical features from FPFH \cite{rusu2009fast} for bootstrapping. These features are effective as initialization for RGB-D point clouds but not discriminative enough for radar (see Section~\ref{sec:ablation}). Moreover, FPFH processes only the 3D coordinates and cannot benefit from additional point cloud features such as color or radar cross-section. After removing the augmentation for the teacher, we found that FPFH features are counterproductive. So, we omitted them and instead trained them with a teacher network that is randomly initialized.

\section{Experiments}
We evaluate our method on the widely used 3DMatch benchmark, a collection of RGB-D video data from indoor scenes. We also test on a proprietary automotive radar dataset collected for a mapping task. Specifically, it was designed to register point clouds from different drives, so each road must have been driven multiple times.
This task is underrepresented in state-of-the-art point cloud registration, as the standard approach for automotive datasets \cite{Geiger2013IJRR,caesar2020nuscenes, sun2020scalability} only registers different frames from the same drive.

\subsection{Baselines}
To provide a more insightful comparison of our approach with existing unsupervised methods, we adhere to the widely used FCGF \cite{choy2019fully} feature extractor as our backbone network. However, \ours\ should be applicable to any other trainable feature matcher, such as GeoTransformer \cite{qin2022geometric} or BUFFER \cite{ao2023buffer}.
We compare \ours\ against the supervised setting \cite{choy2019fully}, SGP \cite{yang2021self}, and EYOC \cite{liu2024extend}, which all train an FCGF model.

For SGP, we partition the total number of epochs into 8 training runs for 3DMatch and 4 runs for the radar data. In each run, the model is further fine-tuned on the new pseudo labels. EYOC utilizes consecutive frames from point cloud sequences for its progressive dataset. Since this is not always available and for a better comparison against all other methods, which cannot access this additional data, we exclude the progressive dataset for our experiments. Additionally, we exclude its spatial filtering, as we have found it inapplicable to RGB-D indoor scenes and to radar scans, which have already undergone post-processing. Note that we use dataset-tuned parameters for its SC$^2$-PCR estimation.

\subsection{Datasets and Implementation}
\noindent\textbf{3DMatch:}
We evaluate on the RGB-D dataset 3DMatch \cite{zeng20173dmatch}. It contains 62 indoor scenes, where the point cloud pairs overlap by at least $\SI{30}{\%}$. We split the data according to the FCGF experiments \cite{choy2019fully} into 48 training, 6 validation, and 8 test scenes. We use the validation set to select the best-performing model in terms of the unsupervised feature match recall (FMR), except for the supervised model validated on the ground truth FMR. It is defined as the percentage of samples with an unsupervised  $\operatorname{IR}(\hat{\transf}, \corrEstNew)$  or ground truth inlier ratio $\operatorname{IR}(\transf^*, \corrEstNew)$ (see Eq.~\ref{eq:inlier}) above $\SI{5}{\%}$ \cite{deng2018ppfnet}. All models are trained for 200 epochs and have seen only the training scenes. Further, we follow the experiment setup of SGP. Thus for all evaluated methods, we use a voxel size of $\SI{5}{cm}$. For inference (and pseudo label generation of SGP and \ours), we apply RANSAC with $\SI{10}{k}$ iterations followed by ICP and measure the registration recall (RR) by calling a registration as successful if the relative translation error (RTE) is below $\SI{30}{cm}$ and the relative rotation error (RRE) is below $\SI{15}{^{\circ}}$.

\noindent\textbf{Automotive Radar:}
The dataset is designed so that each road is driven several times. It was collected from a few cars, each equipped with a long-range consumer-grade radar sensor. The driving was conducted on highways, city streets, and rural roads. Moving objects were filtered out. We split the data into approximate $\SI{250}{k}$ training, $\SI{30}{k}$ validation, and $\SI{30}{k}$ test pairs by their geographic location. Each point cloud contains several hundred points, along with additional features such as radar cross-section.
Given the paucity of information in the z-axis, we remove it and learn 2D point cloud registration instead. We apply RANSAC with $\SI{5}{k}$ iterations without ICP during training and evaluation and use a voxel size of $\SI{50}{cm}$. All methods are trained for 16 epochs.
Analogous to 3DMatch, we report on the registration recall with $50cm$ and $\SI{1}{^\circ}$ as thresholds for RTE and RRE, respectively.

\subsection{Results}

\begin{table}[t]
\begin{center}
\begin{tabular}{lS[table-format=2.1]S[table-format=2.1]S[table-format=2.1]}
\hline
Method & \multicolumn{1}{c}{\begin{tabular}[c]{@{}c@{}}FMR \\ (\%) $\uparrow$\end{tabular}} &  \multicolumn{1}{c}{\begin{tabular}[c]{@{}c@{}}RR \\ (\%) $\uparrow$\end{tabular}} & \multicolumn{1}{c}{\begin{tabular}[c]{@{}c@{}}IR \\ (\%) $\uparrow$\end{tabular}} \\
\hline\hline
Supervised \cite{choy2019fully} & \bfseries 93.5 & \bfseries 92.0 & \bfseries 24.3  \\\hline
SGP \cite{yang2021self} & 91.3 & 90.8  & 22.4
 \\
EYOC \cite{liu2024extend} & 62.5 & 76.8 & 10.6 \\
\ours\ (Ours), $\theta^t = \theta^s$ & 92.3 & 91.1 & 22.4 \\
\ours\ (Ours) & \underline{92.7} & \underline{91.6} & \underline{24.1} \\
\hline
\end{tabular}
\end{center}
\caption{\textbf{Registration results on 3DMatch.} $\theta^t = \theta^s$ means that the student and the teacher network share the same parameters, i.e. no momentum teacher is used.}
\label{tab:3dmatch}
\end{table}

\noindent\textbf{3DMatch:}
The evaluation results are reported in Tab.~\ref{tab:3dmatch}. Our approach yields the best performance among all unsupervised methods on feature match recall, registration recall, and inlier ratio and is only surpassed by the supervised FCGF. A more rigorous version of ours, where the student and teacher share the same parameters (i.e. $\theta^t = \theta^s$) ranks third and can be considered as a more memory-friendly training alternative. SGP achieves comparable results with a marginal decline in performance. It is somewhat surprising that the EYOC method is not able to perform satisfactorily. Therefore, it seems plausible to be caused by the data augmentation applied to the student's and the teacher's input. It is important to note that EYOC was developed for distant point clouds. Consequently, its capabilities are not fully realized when applied indoors.

\begin{table}[t]
\begin{center}
\begin{tabular}{lS[table-format=2.1]S[table-format=1.3]S[table-format=1.3]}
\hline
Method & \multicolumn{1}{c}{\begin{tabular}[c]{@{}c@{}}RR \\ (\%) $\uparrow$\end{tabular}}  & \multicolumn{1}{c}{\begin{tabular}[c]{@{}c@{}}RTE \\ (m) $\downarrow$\end{tabular}} & \multicolumn{1}{c}{\begin{tabular}[c]{@{}c@{}}RRE \\ ($^\circ$) $\downarrow$\end{tabular}} \\
\hline\hline
Supervised \cite{choy2019fully} & \bfseries 96.6 & \bfseries 0.355 & \underline{0.160} \\\hline
SGP \cite{yang2021self} & 90.2 & 0.596 & 0.219\\
EYOC \cite{liu2024extend} & 91.7 & 0.487 & 0.166 \\
\ours\ (Ours), $\theta^t = \theta^s$ & \underline{96.1} & \underline{0.390} & 0.181 \\
\ours\ (Ours) & 95.8 & 0.413 & \bfseries 0.156\\
\hline
\end{tabular}
\end{center}
\caption{\textbf{Registration results on the radar data.}}
\label{tab:radar}
\end{table}

\noindent\textbf{Automotive Radar:}
Tab.~\ref{tab:radar} presents the results for the radar dataset. Once again, our approach's performance is close to that of the supervised method, although the impact of our momentum teacher can be disregarded.
In contrast to the previous experiment, EYOC performs reasonably well, which fits our observation that correct data augmentation is more crucial in 3DMatch. SGP exhibits suboptimal results. We attribute this to the use of learning-free FPFH features and their verifier as we show in ablations in Section~\ref{sec:ablation}. These results underline the generalizability of our method across modalities since no modality-specific hand-crafted features are needed.

\subsection{Ablations}
\label{sec:ablation}

\begin{figure}
    \centering
    \includegraphics[width=0.5\textwidth]{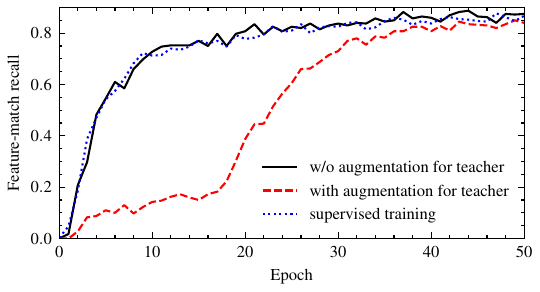}
    \caption{\textbf{Training with and without augmentation for teacher.} Student's feature-match recall on the 3DMatch validation set during training. While the training without augmentation follows the supervised trajectory, the training with augmentation needs more epochs to overcome the bootstrap phase.}
    \label{fig:without_da}
\end{figure}

\noindent\textbf{Data Augmentation for Teacher:}
In Fig.~\ref{fig:without_da}, we demonstrate that the removal of data augmentation for the teacher's input significantly enhances the robustness of the training process. We saw, that keeping the augmentation at best results in a prolonged training process. At worst, the model even fails to converge. After a successful bootstrap phase, it might be advantageous to bring the teacher augmentation back again and then pass different views to the student and the teacher \cite{grill2020bootstrap, caron2021emerging}. We leave this to future work.

\noindent\textbf{Importance of individual components:} Tab.~\ref{tab:ablation} shows an ablation study on a subset containing one-third of the radar scans to evaluate the effect of individual components adopted from SGP. We saw that the additional ICP does not improve the final performance and hence omit it for our radar experiment. 
It is noteworthy that the pseudo label verifier exhibits suboptimal performance, which could be caused by the fact that it removes a substantial number of challenging samples. Furthermore, it may necessitate additional adjustments when utilized for novel datasets.
We also see that a straightforward application of the learning-free FPFH features fails to work on radar and attribute this to the fact that these features are solely based on the 3D coordinates, with the additional radar-specific features not being considered. Both of these observations can be attributed to the suboptimal performance of SGP as seen in Tab.~\ref{tab:radar}.

\begin{table}
\begin{center}
\begin{tabular}{lS[table-format=2.1]S[table-format=1.3]S[table-format=1.3]}
\hline
Method & \multicolumn{1}{c}{\begin{tabular}[c]{@{}c@{}}RR \\ (\%) $\uparrow$\end{tabular}}  & \multicolumn{1}{c}{\begin{tabular}[c]{@{}c@{}}RTE \\ (m) $\downarrow$\end{tabular}} & \multicolumn{1}{c}{\begin{tabular}[c]{@{}c@{}}RRE \\ ($^\circ$) $\downarrow$\end{tabular}}\\
\hline
DiReg with ICP & \underline{95.3} & \bfseries 0.289  & \bfseries 0.093  \\
DiReg w/o~ ICP        & \bfseries 95.7 &  \underline{0.308}  &  \bfseries 0.093  \\
DiReg with Verifier   & 81.6 & 0.935  & 0.141  \\
FPFH + RANSAC         &  3.7  & 52.405 & 62.909 \\
\hline
\end{tabular}
\end{center}
\caption{\textbf{Ablation study} of individual components on a subset of the radar data.}
\label{tab:ablation}
\end{table}

\section{Conclusion}
This study presents a novel self-distillation framework for point cloud registration. We show how to bootstrap student-teacher networks unsupervised without the need for initial hand-crafted features, verifiers, or progressive datasets, while still reaching supervised performance on RGB-D and radar scans.
We hope that future applications can benefit from this work by learning the registration task from large-scale crowdsourced data while needing no ground truth poses and fewer adjustments when dealing with a new modality.

Nevertheless, there is still room for improvement. One possibility might be a non-contrastive loss \cite{grill2020bootstrap} to eliminate the need for negative pairs, as, without ground truth, the pairs are only estimates. Another research direction would be to integrate a differentiable version of RANSAC \cite{wei2023generalized,brachmann2017dsac} and then train in an end-to-end fashion.

\bibliography{egbib}

\end{document}